\documentclass[sigconf]{acmart}

\usepackage[ruled,vlined]{algorithm2e}
\usepackage{multirow}
\usepackage{graphicx}
\usepackage{textcomp}

\AtBeginDocument{%
  \providecommand\BibTeX{{%
    \normalfont B\kern-0.5em{\scshape i\kern-0.25em b}\kern-0.8em\TeX}}}

\setcopyright{acmcopyright}
\copyrightyear{2018}
\acmYear{2018}
\acmDOI{10.1145/1122445.1122456}

\acmConference[CHI LBR '21]{CHI '21: ACM Conference on Computer Human Interaction}{May 08--13, 2021}{Virtual}
\acmBooktitle{Woodstock '18: ACM Symposium on Neural Gaze Detection,
  June 03--05, 2018, Woodstock, NY}
\acmPrice{15.00}
\acmISBN{978-1-4503-XXXX-X/18/06}

\usepackage{color}
\usepackage{ulem}

\copyrightyear{2021}
\acmYear{2021}
\setcopyright{acmlicensed}\acmConference[CHI '21 Extended Abstracts]{CHI Conference on Human Factors in Computing Systems Extended Abstracts}{May 8--13, 2021}{Yokohama, Japan}
\acmBooktitle{CHI Conference on Human Factors in Computing Systems Extended Abstracts (CHI '21 Extended Abstracts), May 8--13, 2021, Yokohama, Japan}
\acmPrice{15.00}
\acmDOI{10.1145/3411763.3451596}
\acmISBN{978-1-4503-8095-9/21/05}

\begin{document}

\title{Developing a Conversational Recommendation System\\for Navigating Limited Options}

\author{Victor S. Bursztyn}
\email{v-bursztyn@u.northwestern.edu}
\affiliation{%
  \institution{Northwestern University}
  \city{Evanston}
  \state{IL}
}

\author{Jennifer Healey}
\email{jehealey@adobe.com}
\affiliation{%
  \institution{Adobe Research}
  \city{San Jose}
  \state{CA}
}
\author{Eunyee Koh}
\email{eunyee@adobe.com}
\affiliation{%
  \institution{Adobe Research}
  \city{San Jose}
  \state{CA}
}
\author{Nedim Lipka}
\email{lipka@adobe.com}
\affiliation{%
  \institution{Adobe Research}
  \city{San Jose}
  \state{CA}
}
\author{Larry Birnbaum}
\email{l-birnbaum@northwestern.edu}
\affiliation{%
  \institution{Northwestern University}
  \city{Evanston}
  \state{IL}
}



\renewcommand{\shortauthors}{Bursztyn and Healey, et al.}



\begin{abstract}
We have developed a conversational recommendation system designed to help users navigate through a set of limited options to find the best choice. Unlike many internet scale systems that use a singular set of search terms and return a ranked list of options from amongst thousands, our system uses multi-turn user dialog to deeply understand the user's preferences.  The system responds in context to the user's specific and immediate feedback to make sequential recommendations.  We envision our system would be highly useful in situations with intrinsic constraints, such as finding the right restaurant within walking distance or the right retail item within a limited inventory.  Our research prototype instantiates the former use case, leveraging real data from  Google Places, Yelp, and Zomato.  We evaluated our system against a similar system that did not incorporate user feedback in a 16 person remote study, generating 64 scenario-based search journeys.  When our recommendation system was successfully triggered, we saw both an increase in efficiency and a higher confidence rating with respect to final user choice.  We also found that users preferred our system (75\%) compared with the baseline.
\end{abstract}





\maketitle

\section{Introduction}
Conversational recommendation systems (CRS) are dialog-based systems that can refine a set of options over multiple turns of a conversation \cite{sun2018conversational, zhang2018towards, christakopoulou2016towards, bridge02towardsconversational, aha1997refining}. Although voice-based assistants are popularizing the use of more conversational interfaces for information-filtering tasks such as shopping online, these approaches to CRS are more focused on parameterizing a search than on getting to know the user during a decision journey. Conventional approaches to recommendation systems (recsys), on the other hand, usually focus on very large sets of options.

The scenario we address is that of finding the best choice amongst a limited set of options. Despite not receiving as much attention, this scenario is actually quite common: finding a restaurant within walking distance, a travel destination under resource constraints, live entertainment, a particular rental car nearby, or an item in stock in a physical store. These nuanced decisions require a deeper understanding of a user's particular needs and preferences, similar to how skilled sales agents get to know a person over multiple rounds of conversation and multiple suggestions, noting sequential user responses and iteratively improving recommendations. For the consumer, the success over their interaction with the agent could be envisioned to be a function of how satisfied they were with their final purchase, how efficient the process was, and how much they enjoyed it. 

We present a system for guided conversational recommendation that is designed to act like a sales agent in a constrained environment as described above. We have implemented a working prototype for the restaurant domain in which users can navigate through real menus, photos, and reviews while considering nearby options. Our contributions are the following: (1) We design our system to actively elicit preferences. (2) We propose a novel language model-based method for preference understanding that does not limit the feature space a priori. (3) We propose a recommendation approach based on extracting arguments from reviews. And (4) we run a small-scale user study assessing two primary goals: confidence upon acceptance, and conversational efficiency. We test the hypothesis that a user would find our guided CRS system more enjoyable, more efficient, and more effective (i.e., more satisfied with the final choice) than using a baseline system that did not sequentially refine the options based on user feedback.

\section{Related work}

There has been an emerging trend in the recsys community towards CRS \cite{sun2018conversational, zhang2018towards}. \cite{christakopoulou2016towards} is one example in the restaurant domain where users are asked to rate specific options. Although it shows benefits of interactive preference elicitation, it doesn't reflect how real conversations unfold. Earlier work traces back to case-based reasoning \cite{bridge02towardsconversational, aha1997refining} and critiquing-based recsys \cite{mahmood2009improving, chen2012critiquing}, which posits that critiquing a choice is a natural way for communicating preferences in real conversations. However, these works generally limit the feature space for expressing preferences. For instance, \cite{chen2017explaining} rely on handcrafted dictionaries that limit the scope of what users can say to only 6-11 features, and \cite{ricci2007acquiring} present a critiquing-based recsys where users can only utilize a predefined set of critiques.

Another trend in information-filtering systems has been focused on multi-modal interfaces, in which pictures are presented to users to help elicit ``unconscious'' preferences in more emotional domains such as tourism and food \cite{neidhardt2014eliciting, jugovac2017interacting}. More recent work has focused on the interplay between product pictures and preferences expressed in natural language \cite{yu2019visual, guo2018dialog}, using deep neural networks for large-scale information-filtering. Preference elicitation has also become more central in recent methodologies for dataset generation \cite{radlinski2019coached, byrne2019taskmaster}, moving away from a predefined feature space---which leads to overly scripted conversations---and closer to human-like conversations.

In this work, we constrain the set of options while keeping the feature space as open as possible. We develop a multi-modal CRS that elicits preferences in a natural way by prompting the user to express open-ended critiques, understanding them, and then selling the user on a choice---which can, in turn, prompt another critique-to-recommendation cycle.

\section{System Description}
\label{methodology}

Our prototype system allows users to navigate through restaurant options nearby, retrieve details including menus, photos, and reviews, express preferences in natural language, and respond to suggestions of restaurants that are potentially appealing, through a multi-modal and conversational interaction\footnote{We use Rasa \cite{bocklisch2017rasa} to build the conversational interface trained on 606 utterance examples handcrafted for this CRS.}. The system makes suggestions by matching one or more user preferences to comments from real customer reviews, which are presented to the user both as an explanation for the recommendation and as an argument aimed at selling them on the specific restaurant.

\begin{figure*}[htb]
  \centering
  \includegraphics[width=.9\textwidth]{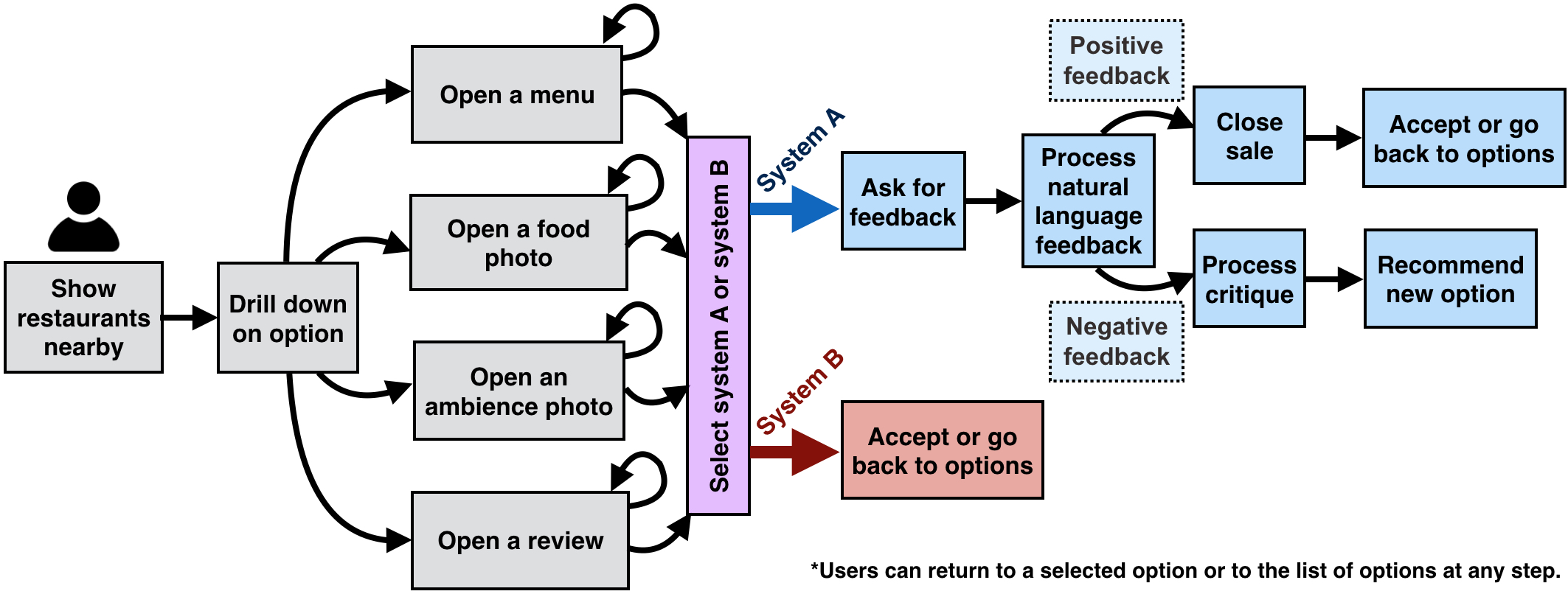}
  \caption{Method overview: The user starts with nearby restaurants, selects one, and interacts with different views (grey). Our system (system A, blue), when used, actively asks for feedback and then processes that feedback; if positive, the system attempts to close the sale; if negative, it takes the critique into account in making another recommendation. The baseline system (system B, red), when used, only reacts to navigation commands.}
  \label{fig:overview}
  \Description{High-level description of the implemented conversational recommendation system.}
\end{figure*}

Figure \ref{fig:overview} provides an overview of our guided CRS, described in more detail below.

\textbf{Navigation actions.} Users start with five restaurant options near their location. As they drill down on one, they can see its price range and cuisine type, or open a more detailed view. We integrate data from Google Places, Yelp, and Zomato, allowing users to navigate through real menus, food and ambience photos, and customer reviews---collectively referred to as ``views.'' Users can always return to their initial list of options, and also refresh it randomly.

\textbf{Active preference elicitation.} If a user clicks on a generic dislike button or walks away from an option after opening a view, our system (system A, blue in Figure \ref{fig:overview}) proactively asks the user what is it about said view from said option that the user did not like.

\textbf{Critique understanding.} If an incoming user feedback is classified as negative by our system\footnote{We provide 243 negative feedback examples and 72 positive ones within our 606 training utterances for Rasa.}, it applies a novel language model-based inference approach to open-ended critique understanding. Specifically, it may be hard for our system to directly apply open-ended negative critiques (e.g., ``That's not good for a date''), since these may not match restaurant attributes as expressed in user reviews; thus we transform these critiques into positive preferences (e.g., ``I prefer more romantic'').

As seen in GPT-3 \cite{brown2020language}, by virtue of their large training corpora, large neural language models are capable of flexible natural language inference if given a few conditioning examples in their prompts. Using this approach, we perform open-ended critique-to-preference transformation by using XLNet \cite{yang2019xlnet} with 7 examples\footnote{A Table included in the supplementary material describes an evaluation for $n \in \{1, 3, 5, 7\}$ examples over a validation set of \textbf{47 unseen critiques}.} in its prompt before appending the critique we would like to transform. Besides being feature-agnostic and therefore more natural than dictionary-based methods, this novel method is also capable of more flexible interpretation of language, such as transforming ``How come they only serve that much?''---with no clearly negative words---into ``I prefer larger portions.''

\textbf{Recommending a new option.} Once our system has a positive preference after active preference elicitation and critique understanding, it's ready to make a recommendation. As detailed in Algorithm \ref{alg:alg1} below, it does this by matching positive preferences to good arguments found in customer reviews. We consider good arguments in favor of a specific choice to be review sentences with both positive sentiment and high alignment (i.e., specificity) w.r.t a preference statement. Hence we parse sentences in user reviews with spaCy \cite{honnibal2017spacy} and, to satisfy the first factor, we filter the ones with a minimum amount of ``joy'' as detected by EmoNet \cite{abdul2017emonet}; then, to satisfy specificity, we represent these filtered sentences in an embedding space using the Universal Sentence Encoder \cite{cer2018universal} and rank them by their cosine similarities w.r.t each (positive) preference statement inferred from user critiques. We then select the positive review sentence that is most aligned as our best argument and recommend the associated restaurant.

\begin{algorithm}[htb]
\small
\SetAlgoLined
\KwResult{The most persuasive sentence (i.e., best argument)}
\texttt{%
 preferences = [``I prefer a good steak",
                ``I prefer a chill place"]\;
 reviews = [review\_1, review\_2, review\_3, review\_4, review\_5]\;
 sentences = parse\_sentences(reviews)\;
 positive\_sents = filter\_positive(sentences, threshold=0.85)\;
 \For{pref in preferences}{
  aligned\_sents = rank\_semantic(pref, positive\_sents)\;
  arguments.add(max(aligned\_sents))\;
 }
 \KwRet max(arguments)\;
}
 \caption{Argument extractor using customer reviews}
 \label{alg:alg1}
\end{algorithm}

\begin{figure*}[htb]
  \centering
  \includegraphics[width=0.9\textwidth]{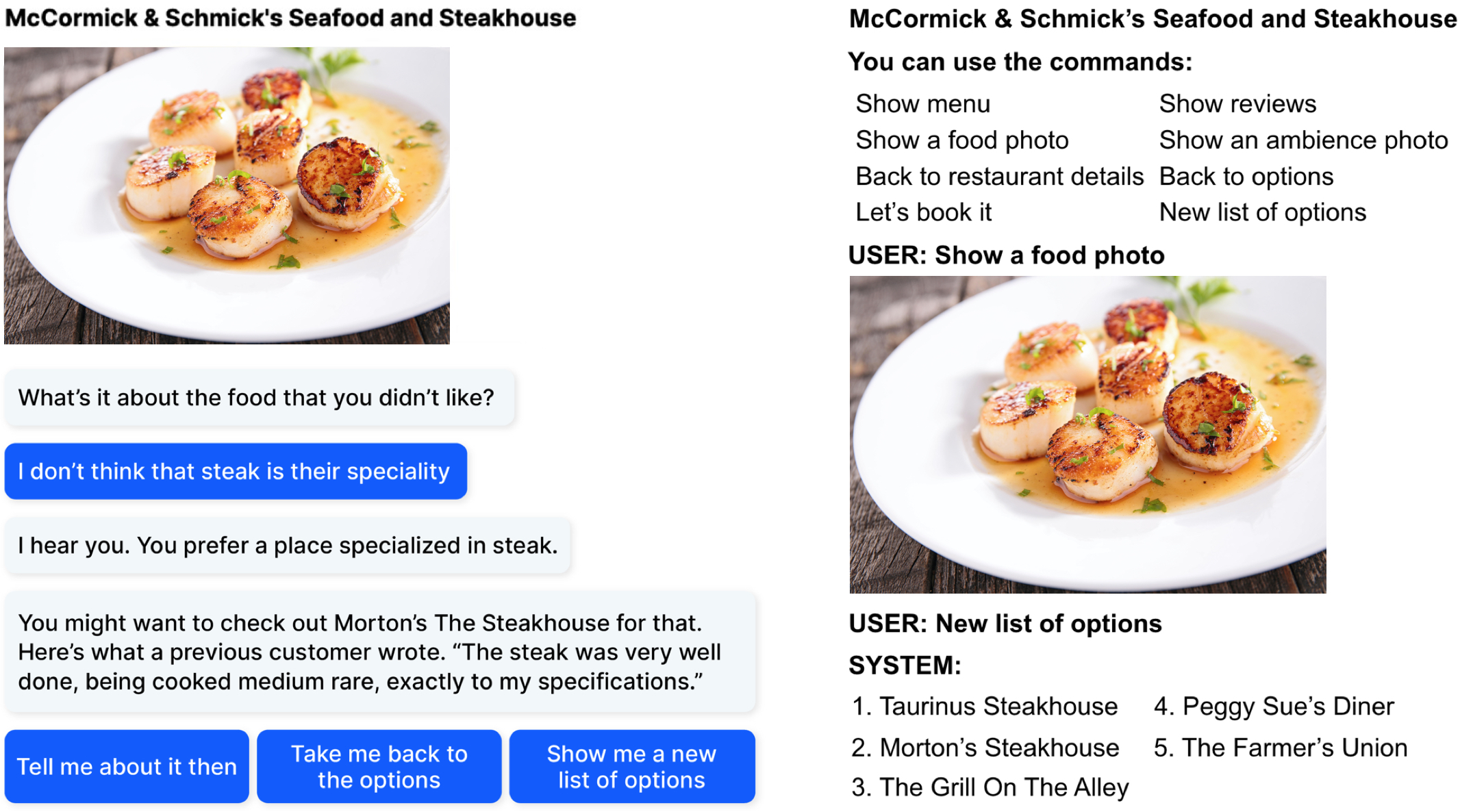}
    \caption{A screenshot of the two systems the user experienced: the guided CRS (left), i.e., system A in Figure \ref{fig:overview}; and the baseline system that did not respond to user feedback (right), i.e., system B.}
  \label{fig:experience_screenshots}
  \Description{Screenshots of the two experiences}

\end{figure*}

Once our system has determined a recommendation supported by an argument, it displays these to the user, who can choose to take it or return to the original options (a real example is seen in Figure \ref{fig:experience_screenshots}, left-hand side).

\textbf{Closing a sale.} If incoming user feedback is classified as positive, the system attempts to close a sale asking if the user would like to book a table. Finally, by contrast, the baseline system (system B, red in Figure \ref{fig:overview}) does not perform active preference elicitation, critique understanding, or recommendations---users need to ask for information until they spontaneously express approval.

\section{User Study}
We performed a small scale user study to test our hypothesis that users would find our active guidance system (System A) more effective than a baseline system that did not respond directly to user feedback and did not build and iteratively improve a model of user preferences, System B.  We measured effectiveness by asking users how confident they felt about their final choice; we measured efficiency by counting the number of commands the user needed to issue to the system to arrive at their final choice.
We recruited 16 participants from a university community (7 male, $\mu$=30.2 years, $\sigma$=6.1 years).  The study was conducted entirely online using video conferencing software. Two experimenters participated in the video call with each of the individual participants.  Both systems ran locally on the hardware of one of the experimenters and the experience was presented to the user as a shared screen.  The experimenter controlling the systems was in charge of both alternating the systems and serving as a perfect speech to text translator for the participant.  The second experimenter was in charge of consenting the participant, presenting the search scenario prompts, and directing the user to fill out the exit questionnaire.     

Participants were asked to conduct four restaurant searches, two using our system and two using a baseline system that only allowed the user to navigate using fixed commands and did not offer guidance, as described above. To motivate the user to navigate towards different kinds of restaurants, we presented them with four different motivating scenarios. These can be described as:
\begin{itemize}

\item {\bf Lunch with Time Constraints (LT)} You are given 90 minutes to find lunch for yourself and return to an important meeting. Find the best place.

\item {\bf Lunch with Kids (LK)} You are looking after your sister's three kids ages 13, 10, and 8. They are really hungry! Find the best place for lunch.

\item {\bf Romantic Dinner (DR)} As a job candidate, the corporation wants to treat you and your significant other to a really nice dinner. They are paying! Find the best place. 

\item {\bf Vegetarian Dinner (DV)} Your manager has just asked you if you can take an important business partner for dinner. They are vegetarian. Find the best place.
\end{itemize}

\vspace{5mm}

System A and System B were alternatively paired with each of the searches.  The order of the searches was also rotated to balance the order of presentation.  In our ideal distribution plan, each of the scenarios would have appeared in the first, second, third and fourth order an equal number of times and the order of the system pairings would be similarly balanced.  There were, however, a few execution problems that led to a slight imbalance.  (In one case a scenario was repeated twice to the same participant once using System A and again using System B.  In this case the second experience was discarded.  In another case there was a repeated ordering that led to a small bias with LK(A), LT(B), DR(A), and DV(B) being repeated twice.)

Participants were told that they were testing two different types of CRS, one that was strictly command-based and one that might occasionally ask the user for feedback.  Participants were encouraged to ``think aloud'' during their search process to help us better understand their thinking as they made their decisions.  Each four-scenario session lasted between thirty minutes and one hour depending on the amount of feedback the participant chose to give during the session. 

The scenarios were designed to motivate users to consider different factors when navigating.  LT was designed to emphasize speed; LK was designed to require a kid-friendly restaurant (food, cost, and ambience); DR removed cost constraints and emphasized finding a place for a ``special'' meal; and DV required balancing both the vegetarian meal constraint with the proper ambience for a business dinner.  Participants were told that all restaurants were open at all times, had tables available, required reservations, and that they should use the phrase ``Book it'' to indicate that they were satisfied with their selection.

\subsection{Interaction Analysis}

Our interaction analysis was designed both to evaluate how many commands users needed to issue until their final decision, and to evaluate how the system performed in a naturalistic interaction with real users.  We chose ``number of commands'' as a metric to evaluate the efficiency of the search, rather than time to decision, to allow users as much time as they wanted to ``talk aloud'' about their decision making process.

The restaurant scenarios varied in difficulty based on the limited number of restaurants that we included in the set of restaurants ``within walking distance.'' The ease of completion from most straightforward to most difficult was: LT, LK, DR, and finally DV, with mean number of commands needed to complete each scenario-based search across both systems being, respectively: 5.3, 5.4, 6.2, and (DV) 9.0. A detailed breakdown of the number of commands used on a per system and per scenario basis can be found in Table \ref{tab:results}. 

\begin{figure*}[htb]
  \centering
  \includegraphics[width=0.4\textwidth]{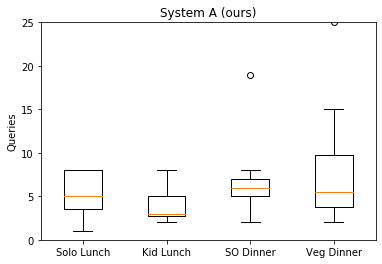}
    \includegraphics[width=0.4\textwidth]{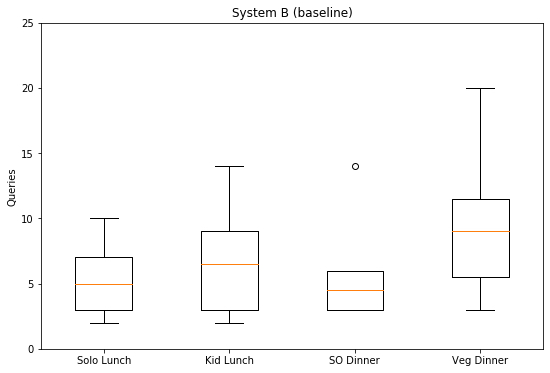}
      \caption{Comparisons of the number of queries users performed using System A (left) and System B (right). In these plots the line shows the median, the box represents the inter-quartile range and the whiskers show the range.  Outliers are represented by dots.}
  \label{fig:boxplots_example}
  \Description{Boxplot data for the interactions}
\end{figure*}

The ``talk aloud'' aspect of the live evaluation exposed a wide variety of assumptions and considerations participants had in mind as they made their decision journeys.  During LT, finding lunch on your own within 90 minutes, we had two nearby restaurants that could easily satisfy this requirement, however one participant chose a more distant restaurant but assumed she could just ask for her order ``to go'' when she got there.  For the lunch with kids scenario, LK, there was one close, low cost restaurant with burgers, but the ambiance photos looked dark.  Some participants chose it anyway while others opted for more expensive restaurants.  The prompt for the romantic dinner, DR, uncovered a number of participant concerns that we had not considered: two participants had partners with gluten-free diets, one had a partner who was vegetarian, and one chose a more casual restaurant, despite the unlimited budget, because it looked like it had dancing. There was no obvious choice for the dinner with a vegetarian business client scenario (DV).  Participants mainly discussed their concerns about what ``type'' of vegetarian the client was.  One participant said ``that would be OK unless they were vegan,'' while another said ``I have known some kinds of vegetarians who will make exceptions for a really nice steak.''  One participant just made his best guess and assumed he could check later with the person if it met their requirements. 
Figure \ref{fig:boxplots_example} shows the variation in the number of commands participants needed to complete each type of search.

\begin{table}
\scalebox{0.85}{
\small
\centering
\begin{tabular}{|l|l|l|l|l|l|l|l|l|l|l|l|}
\hline
\multirow{2}{*}{Scenario}     & \multicolumn{4}{l|}{Commands}                                    & \multicolumn{4}{l|}{Confidence}                               \\ \cline{2-9} 
                              & \multicolumn{2}{l|}{System A}                         & \multicolumn{2}{l|}{System B}                             & \multicolumn{2}{l|}{System A}                            & \multicolumn{2}{l|}{System B}                            \\ \hline
            & $\mu(\sigma)$ &  M   & $\mu(\sigma)$ &  M   & $\mu(\sigma)$ &  M          & $\mu (\sigma)$  &   M         \\ \hline
LT            & 5.3 (\textpm 2.6)&  5          & 5.3 (\textpm 2.5) &5 &  {4.4 (\textpm 0.7)} & 4.5 & 4.1 (\textpm 0.8) & 4        \\ \hline
LK &  3.9 (\textpm 1.9 ) &3&6.9(\textpm 4.1)&6.5 &4.6 (\textpm 0.5)&5& 4.1 (\textpm 0.7) &4\\ \hline
DR & 6.9 (\textpm 4.6) & 6 & \textbf{5.5} (\textpm 3.4) & 4.5 & 4.4 (\textpm 0.7) & 4.5 & 4.3 (\textpm 0.5)& 4 \\ \hline
DV   &  8.5  (\textpm 7.3) & 5.5 & 9.4 (\textpm 5.3) & 9 & 3.2 (\textpm 1.0) & 3 & 3.00 (\textpm 0.9) & 3 \\ \hline
Overall   & 6.2 (\textpm 5.0) & 5 & 6.6 (\textpm 4.1) & 6 &4.1 (\textpm 0.9)  & 4     & 3.9 (\textpm 0.9) & 4  \\ \hline
\end{tabular}}
\caption{Detailed statistics for the performance of System A and System B over the four scenarios in terms of efficiency (number of Commands) and efficacy (user Confidence in the final result). Here the symbol M represents the median value. Confidence is valued on a scale of 1 (very doubtful) to 5 (very confident).}
\label{tab:results}
\end{table}
\normalsize

Apart from these insights, the major finding from our interaction analysis was that our differentiating recommendation process was only successfully activated four times out of the 32 times users experienced our system.  When the system was triggered, it performed quite well.  
In one such instance, the person told the system: {\it ``I didn't see a lot of vegetarian options''} and our system responded {\it ``I hear you. You prefer vegetarian.''} and offered a review of a restaurant that praised the fresh vegetables that it served. Another participant told the system: {\it ``It looks too greasy''} to which the system responded {\it ``I hear you. You prefer lighter food.''} and recommended a choice more to the participant's liking.  Yet another participant commented: {\it ``The ambience was fine but it doesn't seem like a good fit for a business meeting''} to which our system replied {\it ``I hear you. You prefer a more traditional meeting place.''} and offered a suggestion that better met the user's expectations.  

However, due to the sparsity with which our system's unique elements were triggered, the results are highly preliminary.  The triggers occurred three times in response to the lunch with kids scenario (LK), generating command sequences of lengths 8, 3, and 4, and once during the vegetarian dinner (DV) scenario, generating a command sequence of length 4.  This gives some evidence that our system can reduce the number of interactions needed to make the best decision, and that this reduction is most likely to occur for more difficult searches.  This hypothesis is additionally supported in the overall numbers shown in Table \ref{tab:results} where the advantage of our system with respect to efficiency can be most clearly seen in the median results for the most difficult search, DV (5.5 for our system, System A, versus 9 for the baseline system, System B).

\subsection{Questionnaire Analysis}

To assess how confident users felt about their final choice, we asked participants to fill out a short questionnaire after completing all four searches.  The first questions collected age and gender information with the option not to answer.  One participant did not disclose their age.  The second question simply asked which system the user preferred with a forced choice of System A or System B. 75\% participants responded that they preferred our system, System A. 

The majority of the questions were designed to elicit how confident users were about their final choice in each scenario. Users were asked to rate their confidence in their final choice on a scale of 1 to 5 with 1 representing ``Very Doubtful'' and 5 representing ``Very Confident.'' For each participant, the choice of system for each scenario was preassigned and each participant used System A twice and System B twice. The detailed results of the users' confidence in each system broken down by scenario are presented in Table 1.  Across both systems, from most to least confident, the order of the scenarios was LK, DR, LT, and then DV, with respective mean scores of 4.5, 4.4, 4.3, and 3.1. The most visible difference was for the dinner with a vegetarian business client, where there was both no obvious choice and the dining partner was an unknown yet important individual. 
Overall, the users' confidence in both systems was similar, as shown in Table \ref{tab:results}. For scenarios where our agent was successfully activated, our confidence scores were 5, 5, and 4 for the LK scenario, and 3 for the DV scenario, which is slightly higher than the mean score of System B for the LK scenario and in line with the mean score for the DV scenario.  This is inline with our prior discussion of the clearly available options that met the requirements of each scenario.  

\section{Discussion \& Future Work}

In this paper, we presented a guided multi-modal CRS and tested it using a constrained restaurant search application, limited to restaurants within walking distance of a certain location.  We developed a novel language model-based method to infer user preferences from open-ended natural language critiques, and tested in a small-scale user study, similar to \cite{tsai2018beyond} and \cite{tsai2019evaluating}, with highly engaged users to obtain preliminary examples of how the system would behave in response to realistic user search journeys.  Our results showed that our system, when triggered, could respond intelligently to user's feedback, shorten the number of commands necessary to arrive at a decision and potentially boost the user's confidence.  We also found that users preferred our system over the standard command-based system, even though that system should have seemed more familiar.

Our think aloud process allowed us to uncover a number of scenario variables that we plan to address in the future, generally related to stricter settings definitions, e.g., not allowing ``to go'' orders and either asking the user not to consider ``extra-scenario'' dietary restrictions or to account for this in the survey.  We are also considering improving the mechanics of the search by providing, for example, details of gluten free, vegetarian, and kid-friendly options, and showing more than one food or ambience photo at a time.

Our most important consideration for future work, however, is ensuring that our critique-to-recommendation process has a far greater likelihood of being triggered.  To do this, we intend to ask for feedback more explicitly, more often (e.g., when the user asks for a new list of options), and also to focus on longer searches.  We also intend to use the real interaction dialog captured during this initial pilot to further bootstrap our language model-based method.

With respect to improving recommendations, we plan to match inferred preferences not only to customer reviews but also to expanded menu characterizations, e.g., whether the food seems traditional, or eclectic. Additionally, we plan to continue to improve the argument extraction algorithm and also explore negative sentences to directly filter out options. Finally, going beyond our current approach, we will strive to optimize argument extraction---and therefore recommendations---across the entire set of user preferences.

\begin{acks}
This work is supported by gift funding from Adobe Research.
\end{acks}


\bibliographystyle{ACM-Reference-Format}
\bibliography{CHI}
\end{document}